%% file: main.tex
\documentclass[letterpaper, 10 pt, conference]{ieeeconf}  

\IEEEoverridecommandlockouts                              

\usepackage{url}
\usepackage[table,xcdraw]{xcolor}
\usepackage[hidelinks]{hyperref}
\usepackage[utf8]{inputenc}
\usepackage[small]{caption}
\usepackage{amsmath}
\usepackage{amsfonts}
\usepackage{amssymb}
\usepackage{booktabs}
\usepackage{algorithm}
\usepackage{breqn}
\usepackage{multirow}
\usepackage{subcaption}

\usepackage{booktabs}
\usepackage{algorithmicx}
\usepackage{algorithm}
\usepackage[noend]{algpseudocode}
\usepackage{bm}
\usepackage{amsmath}
\DeclareMathOperator*{\argmax}{arg\,max}

\newtheorem{definition}{Definition}

\newtheorem{example}{Example}

\newcommand{\rulesep}{\unskip\ \vrule\ }

\title{\LARGE \bf
Shared Control with Black Box Agents using Oracle Queries
}

\author{Inbal Avraham$^{1}$ and Reuth Mirsky$^{1,2}$
\thanks{        $^{1}$Computer Science Department at Bar Ilan University
                $^{2}$Computer Science Department at Tufts University
        {\tt\small inbala26@gmail.com, reuth.mirsky@tufts.edu}
}}

\newenvironment{myquote}%
  {\list{}{\leftmargin=0.15in\rightmargin=0.15in}\item[]}%
  {\endlist}
  
\begin{document}

\maketitle
\thispagestyle{empty}
\pagestyle{empty}

\begin{abstract}
Shared control problems deal with an agent learning to act in collaboration with other agents or systems. When learning a shared control policy, often a short communication between the agents can significantly reduce running times and improve the system's accuracy. 
We extend the shared control problem to include the ability to directly query a cooperating agent. We consider two types of potential responses to a query, namely oracles: one that assumes that the oracle has a perfect knowledge of the shared system and can provide the learner with the best action it should take, even when that action might be myopically wrong (a teacher), and one with a bounded knowledge limited to its own part of the system (an expert).
Given this additional information channel, this work presents several heuristics for choosing when to query: reinforcement learning-based, utility-based, and entropy-based. These heuristics are aimed at reducing the overall learning cost of a system. Empirical results on two domains show the benefits of querying to learn a better control policy and demonstrate the tradeoffs between the proposed heuristics.
\end{abstract}


\section{Introduction}
\label{sec:intro}
Many modern systems require a human and an artificial agent to work jointly in a coordinated manner. Surgery and service robots \cite{selvaggio2019haptic,wu2015improving}, autonomous vehicles \cite{flad2017cooperative,marcano2020review}, and brain-computer interfaces \cite{philips2007adaptive} all involve combining the actions instructed by a human and an artificial agent to form a shared control system. 
According to Abbink et al.~\cite{abbink2018topology}, 
\begin{myquote}
\textit{``In shared control, human(s) and robot(s) are interacting congruently in a perception-action cycle to perform a dynamic task, that either of them could execute individually under ideal circumstances.''}
\end{myquote}
A key challenge in the design of such a system is that the policy of the human agent is hidden from the artificial agent. The human's hidden policy can be viewed as a \textit{black box}. 
However, existing solutions often assume that the learner can only interact with the black box via execution \cite{dragan2013policy,iosti2020synthesizing}. In many of the mentioned applications, the black box is a cooperative agent, so a simple question can result in great improvements to the learning speed and accuracy of the shared system. 

\begin{figure}[t]
\centering
\includegraphics[trim=0 4.5cm 10cm 0, clip, width=0.48\textwidth]{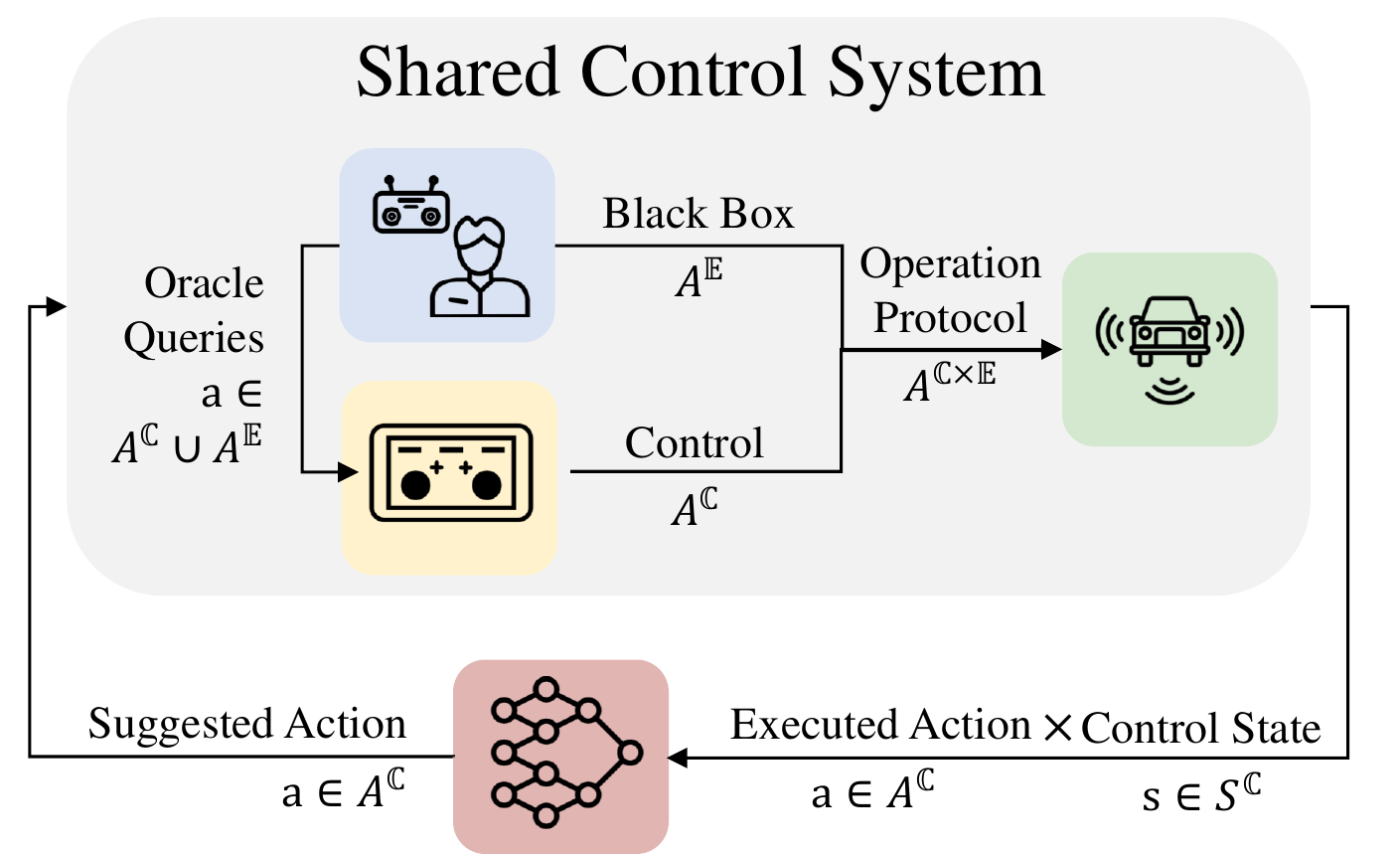}
\caption{\label{fig:system} The new shared system with queries framework.}
\end{figure}

\textbf{The goal of this work is to study the use of queries to improve learning in a shared control system.} This goal is achieved by proposing a new framework for shared control as visualized in Figure \ref{fig:system}. The framework comprises two systems: a black box (representing the human operator) and a control system (representing the autonomous agent). 

In the proposed framework, the control can use the assistance of an \textit{oracle}, which will restrict its potential action space. The oracle may be a cooperating human or an external expert. With the help of such an oracle, learning an efficient policy is expected to be easier, which in turn will minimize the number of failed operations or interactions in the shared system. In this work, we use two oracles types to demonstrate that this expectation is met when the oracle has full knowledge of the shared system, but in other cases, its advice can actually hinder the system's performance. 
Once queries become part of the autonomous agent's optional learning process, a key challenge becomes reasoning about \textit{when} to query the oracle. This paper presents three heuristics that are designed to increase the accuracy of the learned control policy and to minimize the number of queries used during training and execution. 

The contributions of this work are thus threefold: (1) A formal definition of the shared control problem when augmented with queries; (2) An introduction of two oracle types for queries; and (3) Three heuristics for querying in shared control systems.
These contributions are evaluated on two domains: the first is a set of carefully curated shared automata that magnify the unique challenges of learning a shared control policy, and the second is a commonly used shared control simulation of a lunar lander operation \cite{broad2018learning} to show the heuristics' performance in a more natural environment. The results show that the use of queries in those domains can significantly improve learning in both domains (even when the oracle's suggestion might be misleading) and that the different heuristics offer a trade-off between the number of queries used and the accuracy of the learned policy.

\section{Related Work}
Given a shared control system, a key challenge is to design or learn a control that can work with the black box agent to maximize the return of the shared execution.
Dragan and Srinivasa \cite{dragan2013policy} proposed to view the shared control problem as an inverse reinforcement learning problem, in which the task is to elicit the perceived reward of humans from their joint interactions with the control. Alternatively, 
Flad et al. \cite{flad2017cooperative} presented a system that models a human driver's steering motion as a sequence of motion primitives and then uses these primitives to predict the resulting torque. Jiang et al. \cite{jiang2021goal} model a higher-level layer of the intentions of the human rather than their motor primitives. Then, they predict how the shared system should act to accomplish those goals. Iosti et al. \cite{iosti2020synthesizing} learn a shared control for black box environments using a Recurrent Neural Network (RNN). 
The input of the RNN is a vector of the size of the state space, times the action space. The result of the last layer is passed through a softmax function to give the actual output of the network. This output is a vector of a probability distribution for all actions from which the control can select its next step by weighing the probabilities. Thus, once the control chooses to perform an action in a state, and the joint system either accepts or rejects this action, the RNN is updated to reflect the effects of taking that action. 
We use this latter solution as the underlying learner for the control agent, as it is the most unstructured solution, not modeling reward or goal estimations but only on the state and actions.


As often the black box agent in the shared control system is actually a human or a cooperating agent \cite{jain2019probabilistic,javdani2018shared,losey2018review,peternel2018robot}, we wish to leverage an additional communication channel, where the black box assumption is relaxed to allow the control to query the black box about the next best action it should take. 
Querying in such a manner to improve learning has been used in a variety of learning and inference problems, including active machine learning \cite{angluin1987learning,olsson2009literature,settles2012active}, diagnosis \cite{feldman2010model,siddiqi2011sequential}, robotics \cite{abraham2019active, zeng2020view} and planning \cite{brenner2006continual,mirsky2016sequential,shvo2020active}. However, to the best of our knowledge, this work is the first to leverage active learning in shared control systems. We formalize this new communication channel by incorporating an oracle into the shared control system.
\section{The Query-based Shared Control Problem}
\label{sec:definitions}

We start by formalizing the query-based shared control problem using a Multi-Agent Markov Decision Process (MA-MDP) that combines the state and action spaces of the black box agent and the control \cite{albrecht2018autonomous}. The key uniqueness of this model is in splitting the state space into two, according to the visibility of the agents' state, from the learning agent's perspective. This separation of the state space can also be viewed as an extension of Hidden-mode MDP into multiagent settings \cite{choi1999hidden}.
Each of the two spaces can capture an aggregated state of more than one agent in the environment and thus model systems with more than two agents. However, for brevity, in this paper we focus on the simplest case where the hidden part of the state space represents the state of a single black-box agent ($\mathbb{E}$), and the visible part represents the state space of the control ($\mathbb{C}$).

\begin{definition}
A \textbf{Shared System} for control $\mathbb{C}$ and black box $\mathbb{E}$ is defined as a MA-MDP $\langle S^{\mathbb{C}} \times S^{\mathbb{E}}, A^{\mathbb{C}} \times A^{\mathbb{E}}, $T$, $R$  \rangle$, where:
\begin{itemize}
    \item $S^{\mathbb{C}} \times S^{\mathbb{E}}$ is a finite set of states such that $S^{\mathbb{C}}$ are the control's states and $S^{\mathbb{E}}$ are the black box's states.
    \item $A^{\mathbb{C}} \times A^{\mathbb{E}}$ is a finite set of actions such that $A^{\mathbb{C}}$ are the control's actions and $A^{\mathbb{E}}$ are the black box's actions. 
    \item $T: S^{\mathbb{C}} \times S^{\mathbb{E}} \times A^{\mathbb{C}} \times A^{\mathbb{E}} \rightarrow S^{\mathbb{C}} \times S^{\mathbb{E}}$ is the transition of the shared system according to action taken by the control and the black box.
    \item $R: S^{\mathbb{C}} \times S^{\mathbb{E}} \rightarrow \mathbb{R}$ is a function describing the reward of the agents for being at a specific joint state.
\end{itemize}
\end{definition}

 The objective of the shared control can be defined using $R$ to maximize different objectives, such as reaching some goal (as in shared driving, like example \ref{ex:restrict}), maximizing the return (as in atari games, like the lunar lander from Section \ref{sec:results}) or synchronizing the actions of the components (as in a dance performance, or in the automata from Section \ref{sec:automata}). 
In this paper, we keep the transition function deterministic, but it can be extended to stochastic transitions as well. The behavior of the transition function will be determined according to an \textbf{Operation Protocol} (OP), which defines the way that the actions of the agents will be combined. 

\begin{example}
\label{ex:restrict}
Consider a semi-autonomous vehicle (control ${\mathbb{C}}$) that restricts the ability of the driver (black box agent ${\mathbb{E}}$) to change lanes if another vehicle is present in the other lane. A \textit{restriction} operation protocol for the shared system will create a joint action space in which the actions of the driver are $a^{\mathbb{E}} \in \{change,stay\}$, and the actions of the semi-autonomous vehicle are $a^{\mathbb{C}} \in \{allow,restrict\}$. The transition of this shared system is thus
\begin{dmath*}
\begin{aligned}
    & T(s^{\mathbb{E}},s^{\mathbb{C}},a^{\mathbb{E}},allow) = (a^{\mathbb{E}}(s^{\mathbb{E}}), allow(s^V)) \\
    & T(s^{\mathbb{E}},s^{\mathbb{C}},a^{\mathbb{E}},restrict) = (stay(s^{\mathbb{E}}), restrict(s^{\mathbb{C}}))    
\end{aligned}
\end{dmath*}
Intuitively, this transition means that if $V$ allows the driver to change lanes, the driver will be able to execute its transition as expected ($change(s^D)$).
However, the same operation protocol will inhibit the driver's change action if the semi-autonomous vehicle forbids its operation. 
\end{example}


The proposed definition of a shared control system is very broad and can encompass a variety of shared control protocols. Different protocols can let one agent or both manipulate the joint operation or override an action. Alternatively, if the state and action spaces of the control and the black box are the same, a share control system can blend the chosen actions into some weighted sum. 

\begin{definition}
An \textbf{oracle} is a function that gets a joint state and returns a set of actions $O: S^{\mathbb{C}} \times S^{\mathbb{E}} \rightarrow \mathcal{P}(A^{\mathbb{C}} \cup A^{\mathbb{E}})$.
\end{definition}

We do not restrict the answer of the oracle to be an action that is executable by the control or the black box --- the answer depends on the type of oracle used. E.g. a \textit{teacher} oracle is familiar with the system as a whole and can provide the best action(s) the control should take to accomplish the shared objective over the course of an entire execution:
\begin{equation*}
O_{Teacher}(s^{\mathbb{C}},s^{\mathbb{E}})=\{\argmax_{a \in A^{\mathbb{C}}} G(s^{\mathbb{C}}, a)\}    
\end{equation*}
Where $G(s^{\mathbb{C}}, a)$ can lead to the best potential return if the control takes action $a$ in $s^{\mathbb{C}}$. An \textit{expert} oracle is one that is familiar with the black box's behavior and can predict what is the next action that will maximize its reward.

\begin{equation*}
  O_{Expert}(s^{\mathbb{C}},s^{\mathbb{E}})=\{\argmax_{a \in A^{\mathbb{E}}} R(s^{\mathbb{E}}, a)\}      
\end{equation*}


Regardless of the oracle's type, querying can be costly, and hence the control should reason about when to query: during the learning stage only or also during testing, when in a challenging state, and so on. 
The premise of this work is that queries do have some cost, but we do not attribute to this cost a specific value, as such values are domain-dependent. Instead, the aim of the proposed heuristics is to minimize the use of queries, while maintaining high performance.
\section{Query Heuristics for a Shared Control}
\label{sec:heuristics}

To answer the question of when to query the oracle, we propose three different heuristics, each inspired by different factors: The Entropy-based heuristic compares the Entropy from the learned policy's suggestion and the expected Entropy from using the oracle; The Utility heuristic reasons about the probability of getting a successful interaction with a query, conditioned by our prior knowledge about successful transitions that state; The RL-based heuristic takes a Q-learning approach and chooses to query if the expected reward from querying at a state is higher than not querying. 

\subsection{Entropy Heuristic}
\label{sec:IG_heuristic}
The gist of this heuristic is to reason about the confidence of the learner about the information that was collected so far (or whether there's a need to use an oracle). Information Gain (IG) is a measurement that can provide an answer to this question. It is a measure to estimate the confidence level of our learned policy as a predictor for the system's next best action, using entropy estimation. Entropy is defined as the randomness or measuring the disorder of the information in our process \cite{shannon1948mathematical}. It is also a machine learning metric that measures the unpredictability or impurity in the system. When entropy is fairly low, this heuristic provides a useful way to alert for the need for oracle queries. However, if the learned policy (e.g. using the RNN) has high confidence in a wrong action, the Entropy-based metric will only exacerbate this misconception, as seen in the empirical results in Section \ref{sec:results}.
Generally, given a discrete random variable $v$, which takes values in the alphabet $X$ and is distributed according to  $p$:{$X$}\textrightarrow  [0,1], entropy is defined as:
\begin{equation}
{\displaystyle \mathrm {H} (v):=-\sum _{x\in X}p(x)\log p(x)=\mathbf {E} [-\log p(X)]} 
\label{eq:Entropy}
\end{equation}

 The expected information gain from executing $q$ in state $S$ is the difference between the general entropy $\mathrm{H}(S)$ and the entropy given that $q$ was executed ($\mathrm{H}(S\mid q)$). Formally:

\begin{equation}
\mathrm{IG}\left(\mathrm{S,q}\right)\mathrm{=H(S)-H(S \mid q)}
\label{eq:IG}
\end{equation}

When considering the IG with respect to an oracle, $S$ is the state of the joint system and $q$ is the query. Then, $H(S)$ is the entropy when acting in the environment without querying, and similarly  $H(S \mid q)$ is the conditional entropy of $S$ given that we query the oracle.
When the IG is bigger than a certain threshold, the system will use the assistance of an oracle to choose the next step.  

\noindent \textbf{Computing IG}
A challenging part of using the entropy heuristic to decide when to query is computing the IG. Given a shared control system $M$,
 $a^{\mathbb{C}} \in A^{\mathbb{C}}$ the action that was chosen by the control, $a^{\mathbb{E}} \in A^{\mathbb{E}}$ the action that was chosen by the black box, and $a^o \in A^{\mathbb{C}} \cup A^{\mathbb{E}}$ the action that was chosen by the oracle. The IG of the oracle for a state $s \in S^{\mathbb{C}} \times S^{\mathbb{E}}$ is the difference between querying or not:

\begin{equation}
   {IG(s,querying)} = H(s)-H(s|querying)
   {\label{eq:IG_q}}
\end{equation}

\noindent The entropy H($s$) when not using the query is then given by:
\begin{equation}
    {H(s)} = - \displaystyle \sum_{a^{\mathbb{C}} \in A^{\mathbb{C}}}p(s,a^{\mathbb{C}})log(p(s,a^{\mathbb{C}}))
    \label{eq:H(G^s)}
\end{equation}

The value of $p(s,a^{\mathbb{C}})$ is given as the probability of $a^{\mathbb{C}}$ to be the action chosen by the RNN. Before the RNN is trained, these probabilities are initialized to be uniform. 
Going back to Equation \ref{eq:IG_q}, $H(s|querying)$ is harder to compute, as it is the entropy of the control when using the query at state $s$. To compute the overall entropy of the control for all possible oracle responses, we sum this value for all possible $a^o$, as given by equation \ref{eq:H_querying}:

\begin{equation}
    {H(s|querying)} =\displaystyle  \sum_{a^o \in A^{\mathbb{C}} \cup A^{\mathbb{E}}}H(s,a^o)
\label{eq:H_querying}
\end{equation}

Intuitively, $H(s,a^o)$ represents the entropy of the control's choices when using the oracle, and receiving from it action $a^o$ in a state $s$.  To compute it, we sum this value for each chosen action $a^{\mathbb{C}}$, given that the oracle's choice was $a^o$, and multiply it by the probability of using the oracle:

\begin{equation}
\begin{aligned}
H(&s,a^o) = p(s,a^o)H(s|a^o)= \\
& -p(s,a^o) \sum_{a^{\mathbb{C}} \in A^{\mathbb{C}}}p(s,a^{\mathbb{C}}|a^o)log(p(s,a^{\mathbb{C}}|a^o))
\end{aligned}
\label{eq:H_single_state}
\end{equation}


As the probability of $a^o$ is not known in advance, it is estimated using the RNN as in Equation \ref{eq:H(G^s)}. However, since one cannot trust the oracle in states in which it is prone to mislead the control, it can bias the result of this computation to favor the wrong action due to false confidence. To avoid this bias, the next heuristic conditions the use of a query by the prior knowledge about the control's past selections.

To conclude, given a state $s$ and a threshold $\beta_{ent}$, this heuristic chooses to query ($Oracle(s)$) or use its learned action ($Learned(s)$) according to the following condition:

\begin{equation}
    H_{Ent}(s) = \begin{cases}
        Oracle(s), & \text{if } IG(s) > \beta_{ent} \\
        Learned(s),              & \text{otherwise}
    \end{cases}
\end{equation}

\subsection{Utility Heuristic}
\label{sec:util_heuristic}
This heuristic adds a preliminary step before choosing whether to query or not. When reaching a state $s$, it first checks the actions that are available for the control in $s$, and will choose to use the suggestion of the RNN if it gives the action a probability higher than some chosen threshold $\beta_{util}$. 

\begin{equation}
    H_{util}(s) = \begin{cases}
        Learned(s),              & \text{if } p(s, Learned(s)) > \beta_{util} \\
        Oracle(s),               & \text{otherwise}
    \end{cases}
\end{equation}



This heuristic might fail when there is a delayed reward, where the cost of the action chosen at a certain state might only be revealed a few more steps ahead (also known as the credit assignment problem). This type of non-myopic behavior cannot be learned if the control only looks at the immediate utility. To reason about longer horizons, the third heuristic leverages Reinforcement Learning techniques which are aimed at solving such sequential decision-making problems.

\input{three_automata.tex}

\subsection{Reinforcement Learning Heuristic}
\label{sec:RL_heuristic}

Reinforcement Learning (RL) is a type of machine learning in which an agent learns a policy by interacting with the environment and evaluating the outcomes of these interactions. Specifically, an RL problem is a tuple $\langle S, A, R \rangle$ where $S$ is a set of states, $A$ is a set of actions the agent can execute in the environment, and $R: S \times A \rightarrow \mathbb{R}$ is the reward for performing action $a$ at a state $s$ (we do not model the transition function as it is not known to the control). In the shared control learning context, the state space used in this heuristic is the states of the visible control $S^{\mathbb{C}}$, the action space consists of two actions: to query and not to query, and the reward is defined to some positive reward if after a choice whether to query or not to query. 
Notice that this MDP is different from the MA-MDP used to model the shared control system as a whole, where the key difference is that here the decision the agents needs to make is only about querying or not.
The RL heuristic is based on the Q-learning algorithm \cite{sutton2018reinforcement}, where the agent keeps an estimation of the value of executing an action in a state using a value function. This function is called a Q-function and it is defined as $Q: S \rightarrow A$. The heuristic chooses to query if the value of querying is higher than not, according to the Q-function.

\begin{equation}
    H_{RL}(s) = \argmax\limits_{a \in \{Oracle(s), Learned(s)\}}Q(s,a)
\end{equation}

During training, we run an $\epsilon$-greedy version of Q-learning, where the learner selects the action with the highest expected return or a random action at a ratio of $(1-\epsilon) : \epsilon$. 
 In our framework, we define the reward as follows: If the chosen action leads to success, then the reward is 50, and if it fails, the reward is -0.3. In addition, if the last five actions were all successful, an additional reward of 50 is given, thus accumulating to a total reward of 100 for the fifth successful consequent action and onward. This reward was chosen to shape the agent's behavior towards longer success strikes.



\section{Empirical Evaluation}
\label{sec:results}
The underlying learning agent used in this work is the one presented in Iosti et al. \cite{iosti2020synthesizing}. However, as mentioned earlier, it can be replaced by any other learning agent that can process the states and actions of the system and output a recommendation for the next action of the control. The empirical evaluation in this paper consists of two domains: an automata-based system and an Atari-based lunar lander simulation \cite{brockman2016openai}. 

\subsection{Domains}
\subsubsection{Automata-based Shared Control}
\label{sec:automata}

 We follow the definition of an automaton by Iosti et al. \cite{iosti2020synthesizing}. These automata are carefully chosen to challenge the learning of the control policy, due to sparse or delayed rewards.

An \textbf{automaton} A is a tuple $\Omega = \langle S, \iota, A, \delta \rangle$, where $S$ is a finite set of states, $\iota \in S$ it the initial state of the automaton, $A$ is a finite set of actions, and $\delta : (S \times A) \rightarrow S \cup \{\perp \}$ is a partial transition function, where $\perp$ stands for an undefined transition. 
%
For brevity, we use $a(s)$ to define the transition $\delta(a,s)$. Denoting $en(s) = \{a \mid a \in A \wedge \delta(s, a) \neq \perp\}$, i.e., $en(s)$ is the set of actions enabled at the state $s$. We assume that for each $s \in S$, the set of possible actions is not empty, $en(s) \neq \emptyset$.
%
A shared control system 
consists of two automata, a black-box ($\Omega^{\mathbb{E}}$) and the control ($\Omega^{\mathbb{C}}$). 



As this is a black-box environment, the control automaton is visible to the learner, but the black box automaton is not. 
The operation protocol for the two automata is an \textit{agreement} operator, which is defined as:

\begin{dmath*}
\label{eq:transition}
    T(s^{\mathbb{C}},s^{\mathbb{E}},a^{\mathbb{C}},a^{\mathbb{E}}) =
\begin{cases}
    a^{\mathbb{C}}(s^{\mathbb{C}}),a^{\mathbb{C}}(s^{\mathbb{E}}),& \text{if } a^{\mathbb{C}} \in en(s^{\mathbb{E}})\\
    s^{\mathbb{C}},a^{\mathbb{E}}(s^{\mathbb{E}}),              & \text{otherwise}
\end{cases}  
\end{dmath*}

Intuitively, this operation protocol enables both automata to proceed with their chosen action if they are in agreement ($a^{\mathbb{C}}=a^{\mathbb{E}}$). If they are not in agreement, the control stays in its current state, and the black box proceeds with its chosen action regardless of the control. If the automata reach an agreement, then the joint operation is considered successful, otherwise, it is considered a failed operation.
%
%
Three main use cases were used in this evaluation, as they clearly and simply present the challenges of learning control policy \cite{peled2019control}.  


The first use case, Cases (Figure \ref{fig:cases}), is designed with a hidden choice the black box makes that the control cannot learn (when the control is at $c_0$ and the black box is at $e_0$), which leads to a challenge to learn a joint policy with a success rate above $50\%$. When reaching the challenging state, the control can only choose $a$ and will necessarily fail. According to the agreement protocol defined above, the black box necessarily takes a random action at this branch. Later on, the control needs to decide what path to take based on the random decision made before by the black box, but because the decision was random, this decision cannot be learned. This situation means that any information gained from queries during learning cannot be useful during execution. However, this is still an interesting use case to examine the benefit of queries during testing, as an oracle can reveal the hidden selection made by the black box.

The second use case, Strategy (Figure \ref{fig:strategy}), exacerbates this challenge since even after a first successful step, the joint system is designed such that an additional fault must occur before the control can pick an available action. This faulty step means that learning must take into consideration more than one action ahead, causing learning to be harder.

The third use case, Combination-lock (Figure \ref{fig:combination}), is designed such that the control should learn a specific sequence of actions to succeed. In this case, querying can significantly improve training since instead of training the network randomly until it converges, the oracle will directly lead the control to the correct sequence. Moreover, there are no hidden actions as in the previous cases, so if the right action is selected, the learned action will also be useful at runtime. Thus, in this use case, the expectation is that querying will benefit in the learning stage, and will not improve beyond that during the testing stage.

The automata-based system is used for evaluation with the three use cases from Figure \ref{fig:automata}. 
The input layer of the RNN consists of  $|S^{\mathbb{C}}|$ × $|A^{\mathbb{C} }|$ nodes (in our case between 6 to 12) and the output consists of three nodes (the number of the control's actions).


\subsubsection{Lunar Lander Shared Control}
The Atari-based system depicted in Figure \ref{fig:lunar} consists of the Lunar Lander simulation in OpenAI Gym \cite{brockman2016openai}. This domain is a classic rocket trajectory optimization problem that was used in previous work to evaluate shared control algorithms designed to learn from human users' interactions \cite{broad2018learning}. The black box $\mathbb{E}$ is the operator of the lander, and the control $\mathbb{C}$ is its fuel system. The states are shared between the operator and the control, such that $S$ is the current position of the lander (x,y, and angle) and its velocity (x,y, and angular). The operator chooses one of four actions: do nothing, engage the right engine, engage the left engine, and engage the central engine. The control has two actions: inject fuel into the engines or not. The joint operation protocol of this system is the \textit{restriction} protocol, similar to the protocol presented in Example \ref{ex:restrict}: 
\begin{dmath*}
\begin{aligned}
    & T(s^{\mathbb{C}},s^{\mathbb{E}},a^{\mathbb{C}},inject) = (a^{\mathbb{C}}(s^{\mathbb{C}}), inject(s^{\mathbb{E}})) \\
    & T(s^{\mathbb{C}},s^{\mathbb{E}},a^{\mathbb{C}},\neg {inject}) = (nothing(s^{\mathbb{C}}), \neg {inject} (s^{\mathbb{E}}))    
\end{aligned}
\end{dmath*}

Intuitively, the engines will activate only if the operator chooses to engage one of the engines and the control flows fuel. In this domain, the states of the black box and the control are the same, so the teacher and the expert oracle always agree. As such, we only show results using the teacher oracle.
This domain is meant to test the generalization of the proposed heuristics to more realistic domains. It is simpler yet bigger than the automata use case: its state space consists of $5^6$ states, making exploration more challenging. 

\input{queryFig_Combination.tex}

\subsection{Setup} 
\label{sec:setup}
In both domains, for each combination of a scenario, oracle type, and a heuristic, an RNN was trained for $40$ epochs. In the automaton domain, each epoch runs on a gradually increasing number of steps, from one to $19$ steps, and in the Lunar Lander domain, it was trained for $200$ steps in each epoch, which is the length of an episode in the simulation. Each configuration was run five times during training, and five more times during testing. 
After some manual tuning, the heuristics thresholds are set to $\beta_{ent}=0.25$ and $\beta_{util}=0.95$. The hyperparameter values of the Q-learning approach were set to be $\alpha=0.5, \gamma=0.9, \epsilon=0.05$.

\noindent \textbf{Metrics} Two main constructs are evaluated in this work: the \textit{accuracy} of the shared control with each heuristic type, and the total \textit{number of queries} used. The fault rate of the heuristics is compared against several baselines: the RNN learns without queries (``No Oracle''), the control takes a random action (``Random''), the oracle is always being called during training only (``Always Train''), and the oracle is always being called both in training and in test time (``Always Train+Test''). The ``No Oracle'' and ``Random'' provide a lower bound for the shared control's performance, while the ``Always'' querying variants provide an upper bound to how much queries can help with an optimal oracle. When we have a Teacher oracle, ``Always'' will provide us with the best possible action to take in terms of the expected return.

\section{Results}
\label{sec:results}

\input{Fault.tex}

\begin{table*}[h]

\caption{Results of failure rate, number of oracle uses, and total reward in the Lunar Lander domain with and without using heuristics. The results using the heuristics are shown in the three rightmost columns.}
\label{tab: Lunar all heuristics}
\centering
\footnotesize

\begin{tabular}{|l|r|r|r|r||r|r|r|}
\hline
\multicolumn{1}{|c|}{} & \multicolumn{1}{c|}{No Oracle}         & \multicolumn{1}{c|}{Random}          & \multicolumn{1}{c|}{\begin{tabular}[c]{@{}c@{}}Always\\ Train\end{tabular}} & \multicolumn{1}{c||}{\begin{tabular}[c]{@{}c@{}}Always\\ Train+Test\end{tabular}} & \multicolumn{1}{c|}{Entropy}          & \multicolumn{1}{c|}{Util}            & \multicolumn{1}{c|}{RL}               \\ \hline
Failure rate           & \multicolumn{1}{r|}{34.3 $\pm$  0.763} & \multicolumn{1}{r|}{50.5 $\pm$ 1.08} & \multicolumn{1}{r|}{21.86 $\pm$  6.93}                                      & \multicolumn{1}{r||}{0 $\pm$ 0}                                                   & \multicolumn{1}{r|}{27.49 $\pm$ 10.7} & \multicolumn{1}{r|}{23.9 $\pm$ 10.9} & \multicolumn{1}{r|}{28.6 $\pm$ 12.01} \\ \hline
Oracle use (len=200)   &{0 $\pm$ 0}                                  & {0 $\pm$ 0}                               & {200 $\pm$ 0}                                                                      & {200 $\pm$ 0}                                                                           & 17.79 $\pm$ 10.37                     & 19.7$\pm$ 8.55                       & 189.7 $\pm$ 1.4                       \\ \hline
Total reward           & 212.3 $\pm$ 14.92                      & -123.6$\pm$ 20.3                     & 205.13 $\pm$ 6.44                                                           & 213.106 $\pm$ 5.514                                                              & 206.24$\pm$ 3.4                       & 208.15$\pm$ 6.42                     & 207.6 $\pm$ 2.78                      \\ \hline
\end{tabular}
\end{table*}

We will first present accuracy results, then oracle usage and we will end this section with a discussion on important results.
In both Table \ref{tab:faults} and Table \ref{tab:  Lunar all heuristics}, the left columns show the failure rate of the baselines, and the right columns show the failure rate of the heuristics: Entropy, Util, and RL.
For Table \ref{tab:faults}, we show two variants of the RL heuristic
(``Train'' and ``Train+Test''), as there was a significant difference between these conditions, especially in the Strategy use case. 

We first look at Table \ref{tab:faults}, and discuss the baselines. Given the teacher oracle, the two baselines ``Always Train'' and``Always Train+Test'' improve the accuracy of the results in two out of the three example scenarios (``strategy'' and ``combination lock'').
 However, for ``cases'' these baselines did not achieve better performance compared to not using any query (``No Oracle''). 
When using expert oracle in ``strategy'', the failure rate of the ``always train + test'' baseline increases significantly, and 
%
in ``Combination Lock'' the failure rate of the ``Always'' baselines using the expert oracle is very high (approximately 86\%) since the environment can accept two actions, one of which leads to a ``sink'' state. 

So far we talked about our baselines. We now move on to discuss the heuristic. As seen in the table, the Entropy heuristic is able to reach a similar performance to the ``Always'' variants in most cases. As we will show later, it does so by querying much less, thus being more efficient. 

We wish to focus on one specific scenario, to discuss the reason that the Entropy heuristic did not improve accuracy. In ``Combination Lock'', the failure rate with the teacher oracle is very high compared to ``Always''. We checked our failure results over the five tests, and the results are on the extreme, getting either a $0$ or a $98.5\%$ failure rate. The reason for this dichotomy is that a failure happens when the choice at the second step of the RNN at the combination lock can either leads us to a ``sink'' state (See $s_3$ in Figure \ref{fig:combination}) or to a successful continuation of the shared operation. These extreme cases are what lead to this highly variant result.
Util heuristic solves the problem of choosing actions that lead to a sink-state in this use case. The teacher oracle suggests choosing 'a','b', 'a', so the agent does not fall into the sink state and learns better.
Thanks to this change, the failure rate in the Util heuristic is $0.39 \pm 0.88 $ compared to $39.59 \pm 53.77$ in the Entropy heuristic. 

The most interesting result for the RL heuristic is in the ``Strategy'' use case, when using the ``Train+Test'' variant. The failure rate decreases from 52.55 $\pm$ 5.44 to 8.3 $\pm$ 17.92 with the teacher oracle. This improvement is due to the choice of the control to use a query when reaching state $g_2$ (see Figure \ref{fig:strategy}), which is a state that requires a single failure if we want the joint system to succeed at the rest of the trial. This policy means that the RL heuristic makes the system query at state $g_2$, even though it decided not to query in previous heuristics, and then often took a wrong (although myopically correct) action. 

In Table \ref{tab: Lunar all heuristics} we show the results of using the heuristics in the Lunar Lander domain compared to the baselines (as in Table \ref{tab:faults}. 
From the table, it can be seen that the use of heuristics was beneficial in all cases. The failure rate is much lower than not using Oracle or choosing a random action and is very similar to ``Always Train+Test'', where the system always uses Oracle. Additionally, in terms of the average number of queries used decreases to around 18  when using the Entropy or Util heuristics, versus 200 $\pm$ 0 for these ``Always'' baselines. This result means that the system uses the oracle much less and reaches very similar results to always query. Also, the reward when using the heuristics is very similar to the one they receive when always using the oracle. Next, follow with the meaning of these results.

\section{Discussion}

\noindent \textbf{Querying during training.} All heuristics show similar performance compared to the best possible strategy in Cases and Strategy (``Always'' with the teacher oracle), and both the Utility and the RL heuristics exhibit close to the best possible performance in the Combination Lock scenario.

\noindent \textbf{Querying during test time.} The Entropy and Utility heuristics do not seem to leverage the ability to query during test time, even when this ability can improve the performance significantly (In Strategy and Combination Lock with the teacher oracle). The RL heuristic is able to learn to query at the right times to benefit from querying, even though the Q-function does not continue to change during test time.

\noindent \textbf{Learning when not to query.} The expert oracle has the potential to actually hurt the shared control (in the Strategy use case, when always querying in train+test), and hence a good heuristic should learn not just when to query, but also when not to rely on the oracle. All heuristics learned when not to use the oracle, showing similar failure rates in all scenarios.

\noindent \textbf{Number of queries used.} 
The test in Figure \ref{fig:combination_queries} examines whether the RNN decreases its reliance on queries with time. The x-axis shows the number of epochs learned so far, and the y-axis shows the number of queries the control initiated. For each query policy-oracle combination, five tests were held so there are 10 lines in each graph in total. The trends are similar in all three setups, so we focus here only on the combination lock use case. Here, the use of a teacher and an expert oracle are quite distinct from one another. With the teacher, the RNN is often able to learn a policy in which a small number of queries are sufficient, while with the expert, the control keeps querying it more often.


\noindent \textbf{Generalization.} We use the Lunar Lander use case to evaluate how the heuristics generalize to a more realistic environment. Table \ref{tab: Lunar all heuristics} shows that all of the trends observed in the automata use case remain consistent: the failure rate when using queries decreases significantly compared to the same runs without queries. The total number of queries is still the highest in the RL heuristic. The new interesting information from this domain is that the reward, which is an external metric for the performance of the shared system (unlike failure rate which is an intrinsic metric) is not much different between the policy without queries and with them.
\section{Conclusions}
This paper introduces a new formalization for a shared control system with queries. It explores the new formalization by discussing and evaluating two potential oracles that can be queried, as well as three heuristic approaches for choosing when to query in such a shared system.

The results show a clear trade-off between the proposed heuristics: the utility heuristic showed the best results regarding query efficiency, reducing the number of required queries, sometimes to the minimum number of queries in which the true best action cannot be learned. On the other hand, the RL heuristic can reach much lower failure rates but at the cost of a large and inconsistent number of queries. 
These results are consistent across two very different evaluation domains.
Future work will investigate using RL techniques to make querying more efficient and consistent. One option is to replace the RNN itself with an RL algorithm that can choose between the system's different actions and querying. 

Our empirical evaluation also demonstrated the impact of the oracle quality, showcasing its importance. Future work can reason about the oracle type in real time and adapt the chosen heuristic to that oracle. Lastly, another question to investigate is how to design a querying heuristic that accounts for an oracle that intentionally misleads the control.

\bibliographystyle{plain}
\bibliography{bibliography}

\end{document}

%% file: three_automata.tex
\begin{figure*}[t]
     \centering
     \begin{subfigure}[b]{0.15\textwidth}
         \centering
         \includegraphics[trim=3cm 20cm 14cm 2cm, clip, width=\textwidth]{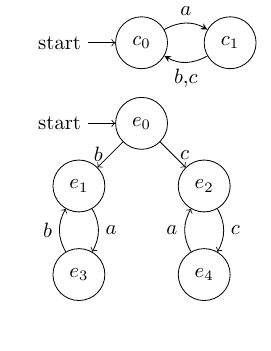}
         \caption{Cases.}
         \label{fig:cases}
     \end{subfigure}
     \hfill
     \rulesep
     \begin{subfigure}[b]{0.20\textwidth}
         \centering
         \includegraphics[trim=3cm 20cm 10cm 2cm, clip, width=\textwidth]{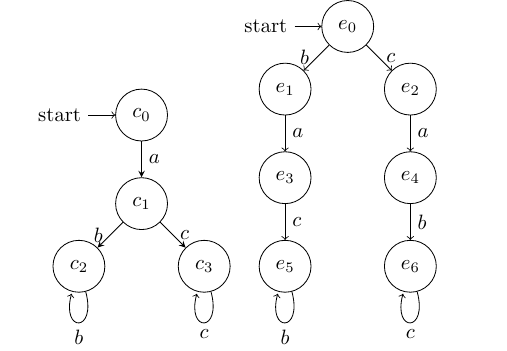}
         \caption{Strategy.}
         \label{fig:strategy}
     \end{subfigure}
     \hfill
     \rulesep
     \begin{subfigure}[b]{0.25\textwidth}
         \centering
         \includegraphics[trim=3cm 19cm 10cm 2cm, clip, width=\textwidth]{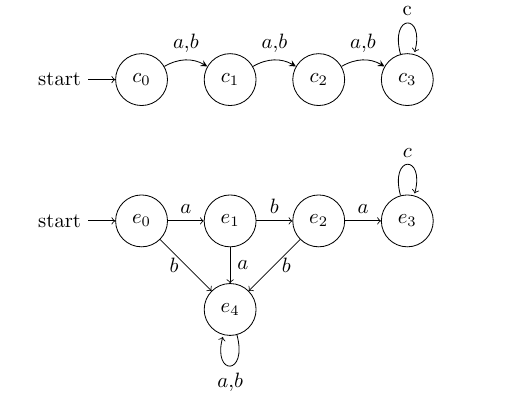}
         \caption{Combination Lock.}
         \label{fig:combination}
     \end{subfigure}
     \hfill
     \rulesep
     \begin{subfigure}[b]{0.25\textwidth}
          \centering
         \includegraphics[trim=0 1cm 0 1cm, clip, width=\textwidth]{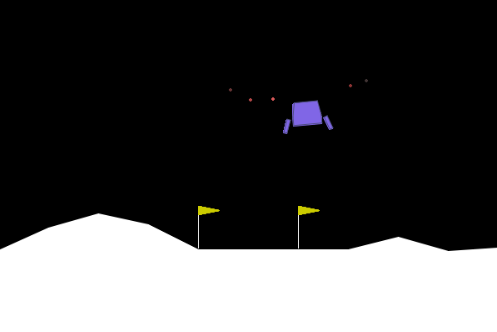}
         \caption{Lunar Lander.}
         \label{fig:lunar}
     \end{subfigure}
        \caption{The use cases from the Automata domain and the Lunar Lander domain, used for the evaluation of the shared control synthesis with queries. For each Automata use case, the states of the control automaton are labeled $c_i$ and the states of the environment automaton are labeled $e_i$.}
        \label{fig:automata}
        \vspace{-2mm}
\end{figure*}

%% file: queryFig_Combination.tex
\begin{figure*}[t]
     \centering
     \begin{subfigure}[b]{0.3\textwidth}
         \centering
         \includegraphics[width=\textwidth]{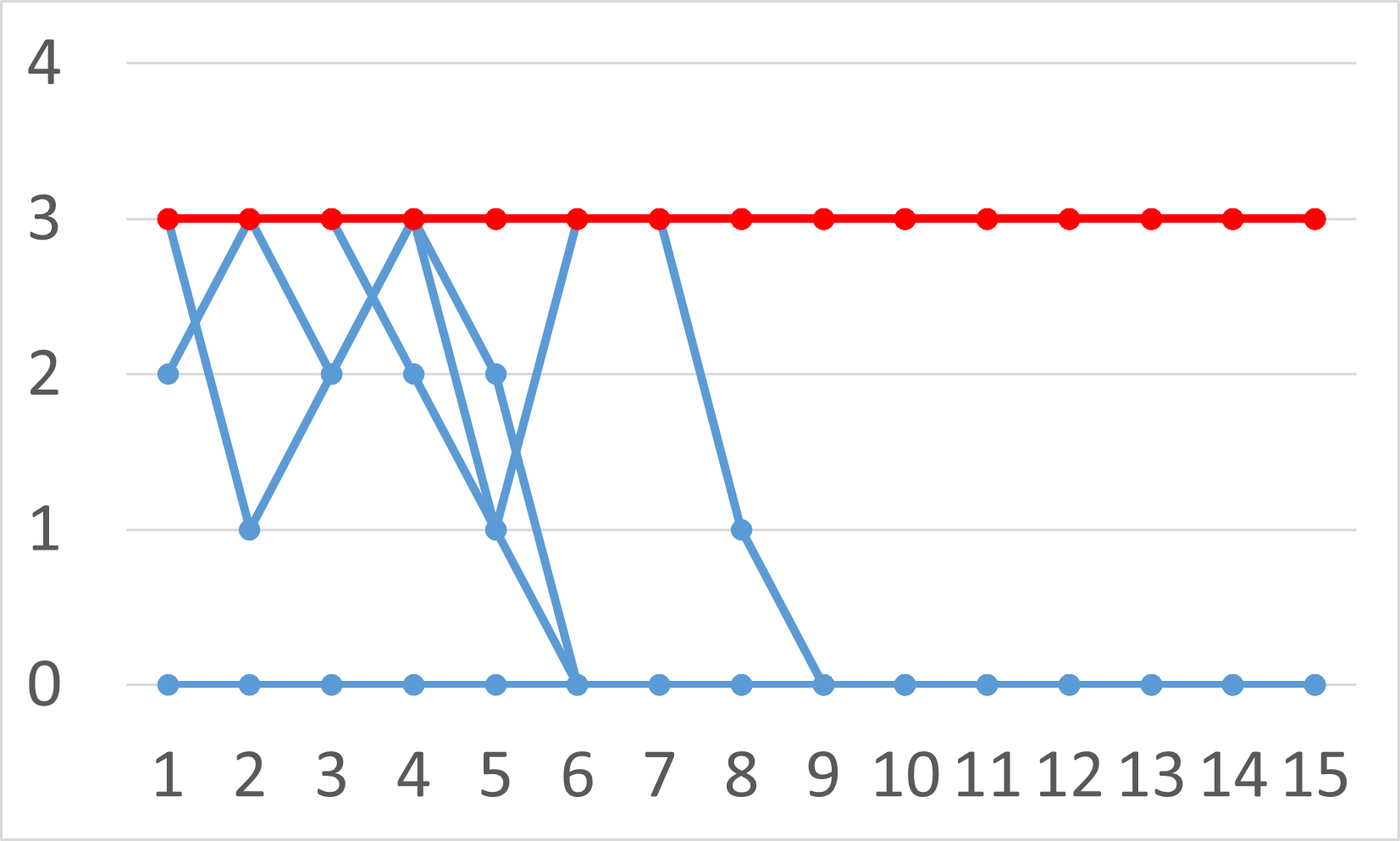}
         \caption{Entropy Heuristic.}
         \label{fig:IG_combination}
        \end{subfigure}
     \hfill
     \begin{subfigure}[b]{0.3\textwidth}
         \centering
         \includegraphics[width=\textwidth]{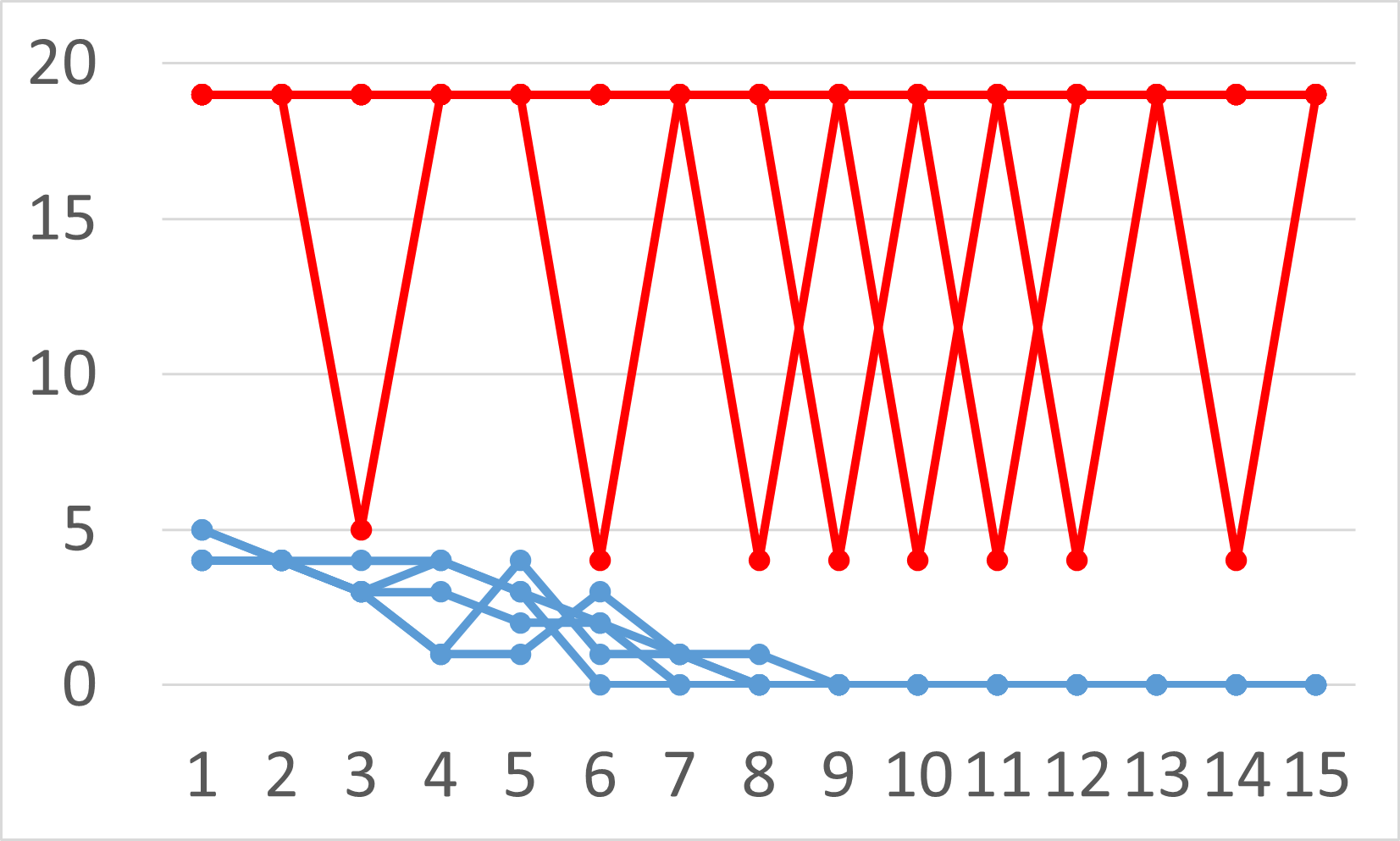}
         \caption{Utility Heuristic.}
         \label{fig:Util_combination}
     \end{subfigure}
     \hfill
     \begin{subfigure}[b]{0.3\textwidth}
         \centering
         \includegraphics[width=\textwidth]{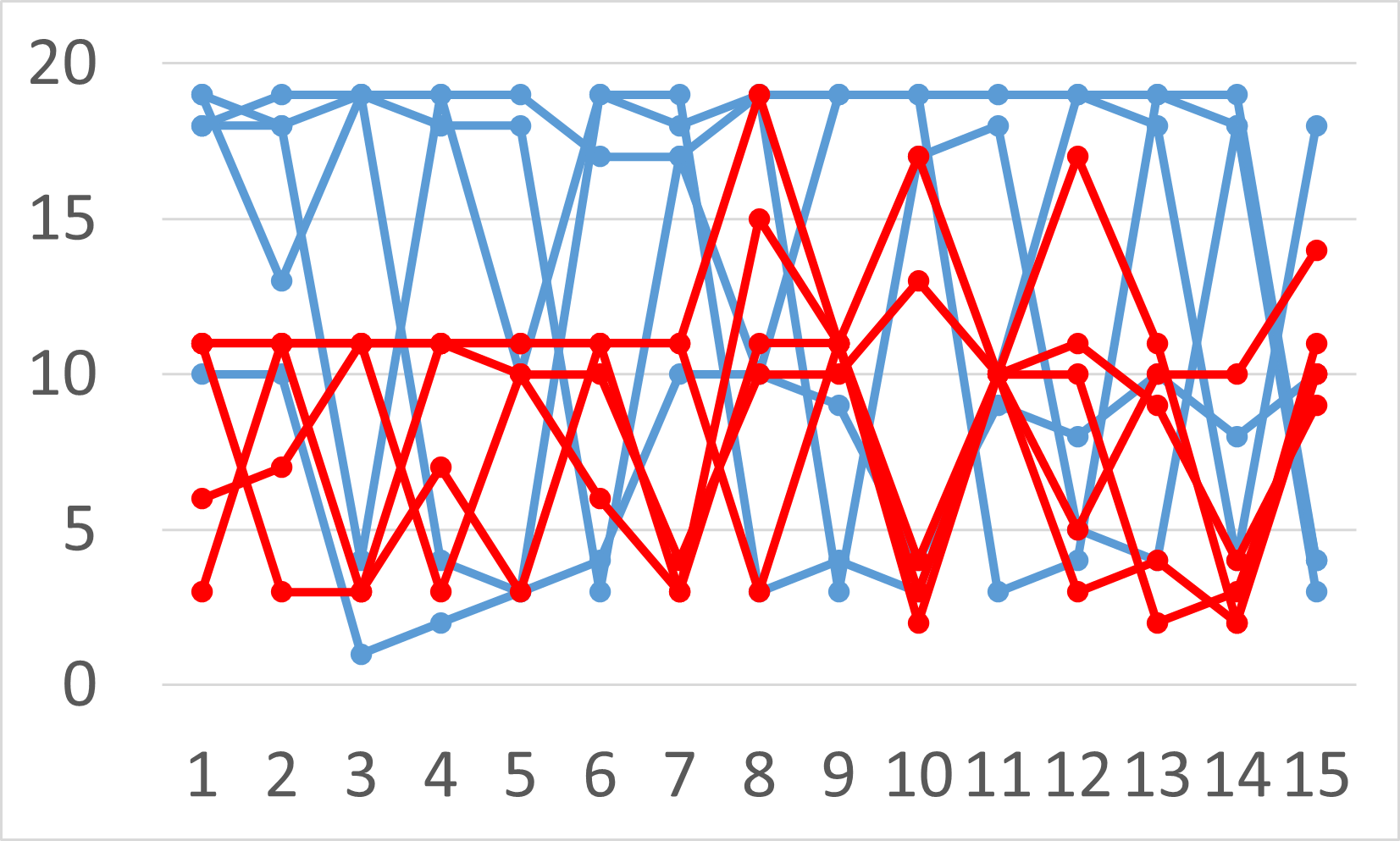}
         \caption{RL Heuristic.}
         \label{fig:RL_combination}
     \end{subfigure}
        \caption{Number of Queries used per training episode of the Combination Lock use case. {\color{red}Red} lines represent tests with the expert oracle, while {\color{blue}blue} lines represent tests with the teacher oracle. The results are cut after 15 episodes, as the algorithm's behavior remains the same.}
        \label{fig:combination_queries}
        \vspace{-4mm}
\end{figure*}

%% file: Fault.tex
\begin{table*}[t]
\setlength{\tabcolsep}{0.5em} 
\caption{Average failure \% (lower is better) across different configurations, including Oracle Type (O) being either Teacher (T) or Expert (E). Baseline runs include ``No Oracle'': the RNN learns without queries, ``Random'': the control chooses a random action, ``Always Train'': the control always uses the Oracle during Training, ``Always Train+Test'': the control always uses the Oracle both in training and in test time. Heuristic runs include ``Entropy'' and ``Util'' are the respective heuristics with queries enabled both in training and test time, and ``RL Train'' and ``RL Train+Test'' for the RL heuristic with queries enabled in training only and in train+test, respectively. }
\label{tab:faults}
\centering
\tiny
\begin{tabular}{ll|rrrr||rrrr|}
\cline{3-10}
                                                &                                                                            & \multicolumn{4}{c|}{Baselines}                                                                                                                                                                                                             & \multicolumn{4}{c|}{Heuristics}                                                                                                                                                                 \\ \hline
\multicolumn{1}{|c|}{Case}                  & \multicolumn{1}{c|}{\begin{tabular}[c]{@{}c@{}}O\end{tabular}} & \multicolumn{1}{c|}{No Oracle}      & \multicolumn{1}{c|}{Random}         & \multicolumn{1}{c|}{\begin{tabular}[c]{@{}c@{}}Always\\ Train\end{tabular}} & \multicolumn{1}{c|}{\begin{tabular}[c]{@{}c@{}}Always\\ Train+Test\end{tabular}} & \multicolumn{1}{c|}{Entropy}         & \multicolumn{1}{c|}{Util}           & \multicolumn{1}{c|}{RL Train}       & \multicolumn{1}{c|}{\begin{tabular}[c]{@{}c@{}}RL\\ Train+Test\end{tabular}} \\ \hline
\multicolumn{1}{|l|}{\multirow{2}{*}{Cases}}    & T                                                                    & \multicolumn{1}{r|}{0.77 $\pm$ 0.04}  & \multicolumn{1}{r|}{50.22 $\pm$ 0.46} & \multicolumn{1}{r|}{1.04 $\pm$ 0.6}                                           & 0.76 $\pm$ 0.03                                                                    & \multicolumn{1}{r|}{0.74 $\pm$ 0.01}   & \multicolumn{1}{r|}{0.77 $\pm$ 0.04}  & \multicolumn{1}{r|}{1.03 $\pm$ 0.61}  & 0.78 $\pm$ 0.06                                                                \\ \cline{2-10} 
\multicolumn{1}{|l|}{}                          & E                                                                     & \multicolumn{1}{r|}{0.8 $\pm$ 0.02}   & \multicolumn{1}{r|}{50.18 $\pm$ 0.56} & \multicolumn{1}{r|}{0.77 $\pm$ 0.04}                                          & 0.77 $\pm$ 0.04                                                                    & \multicolumn{1}{r|}{0.76 $\pm$ 0.02}   & \multicolumn{1}{r|}{0.76 $\pm$ 0.02}  & \multicolumn{1}{r|}{2.64 $\pm$ 7.54}  & 0.87 $\pm$ 0.26                                                                \\ \hline
\multicolumn{1}{|l|}{\multirow{2}{*}{Strategy}} & T                                                                    & \multicolumn{1}{r|}{47.06 $\pm$ 2.19} & \multicolumn{1}{r|}{48.29 $\pm$ 4.4}  & \multicolumn{1}{r|}{53.33 $\pm$ 4.3}                                          & 1.0 $\pm$ 0.0                                                                      & \multicolumn{1}{r|}{51.76 $\pm$ 3.63}  & \multicolumn{1}{r|}{52.55 $\pm$ 5.44} & \multicolumn{1}{r|}{50.2 $\pm$ 5.67}  & 8.3 $\pm$ 17.92                                                                \\ \cline{2-10} 
\multicolumn{1}{|l|}{}                          & E                                                                     & \multicolumn{1}{r|}{48.83 $\pm$ 6.69} & \multicolumn{1}{r|}{49.9 $\pm$ 3.03}  & \multicolumn{1}{r|}{54.72 $\pm$ 6.17}                                         & 99.0 $\pm$ 0.0                                                                     & \multicolumn{1}{r|}{51.41 $\pm$ 3.16}  & \multicolumn{1}{r|}{49.07 $\pm$ 5.47} & \multicolumn{1}{r|}{50.30 $\pm$ 4.55} & 55.98 $\pm$ 19.06                                                              \\ \hline
\multicolumn{1}{|l|}{\multirow{2}{*}{Comb.}}    & T                                                                    & \multicolumn{1}{r|}{98.3 $\pm$ 0.44}  & \multicolumn{1}{r|}{86.88 $\pm$ 2.91} & \multicolumn{1}{r|}{0.2 $\pm$ 0.44}                                           & 0.0 $\pm$ 0.0                                                                      & \multicolumn{1}{r|}{39.59 $\pm$ 53.77} & \multicolumn{1}{r|}{0.99 $\pm$ 0.7}   & \multicolumn{1}{r|}{9.01 $\pm$ 6.22}  & 2.96 $\pm$ 4.55                                                                \\ \cline{2-10} 
\multicolumn{1}{|l|}{}                          & E                                                                     & \multicolumn{1}{r|}{98.5 $\pm$ 0}     & \multicolumn{1}{r|}{86.68 $\pm$ 2.87} & \multicolumn{1}{r|}{87.07 $\pm$ 3.45}                                         & 86.48 $\pm$ 5.04                                                                   & \multicolumn{1}{r|}{86.28 $\pm$ 4.1}   & \multicolumn{1}{r|}{86.47 $\pm$ 5.09} & \multicolumn{1}{r|}{88.45 $\pm$ 5.29} & 86.98 $\pm$ 6.41                                                               \\ \hline
\end{tabular} 
\end{table*}

%% file: main.bbl
\begin{thebibliography}{10}

\bibitem{abbink2018topology}
D.~A. Abbink, T.~Carlson, M.~Mulder, J.~C. De~Winter, F.~Aminravan, T.~L. Gibo,
  and E.~R. Boer.
\newblock A topology of shared control systems—finding common ground in
  diversity.
\newblock {\em IEEE Transactions on Human-Machine Systems}, 48(5):509--525,
  2018.

\bibitem{abraham2019active}
I.~Abraham and T.~D. Murphey.
\newblock Active learning of dynamics for data-driven control using koopman
  operators.
\newblock {\em IEEE Transactions on Robotics}, 35(5):1071--1083, 2019.

\bibitem{albrecht2018autonomous}
S.~V. Albrecht and P.~Stone.
\newblock Autonomous agents modelling other agents: A comprehensive survey and
  open problems.
\newblock {\em Artificial Intelligence}, 258:66--95, 2018.

\bibitem{angluin1987learning}
D.~Angluin.
\newblock Learning regular sets from queries and counterexamples.
\newblock {\em Information and computation}, 75(2):87--106, 1987.

\bibitem{brenner2006continual}
M.~Brenner and B.~Nebel.
\newblock Continual planning and acting in dynamic multiagent environments.
\newblock In {\em Proceedings of the 2006 international symposium on Practical
  cognitive agents and robots}, pages 15--26, 2006.

\bibitem{broad2018learning}
A.~Broad, T.~Murphey, and B.~Argall.
\newblock Learning models for shared control of human-machine systems with
  unknown dynamics.
\newblock {\em arXiv preprint arXiv:1808.08268}, 2018.

\bibitem{brockman2016openai}
G.~Brockman, V.~Cheung, L.~Pettersson, J.~Schneider, J.~Schulman, J.~Tang, and
  W.~Zaremba.
\newblock Openai gym.
\newblock {\em arXiv preprint arXiv:1606.01540}, 2016.

\bibitem{choi1999hidden}
S.~P. Choi, D.-Y. Yeung, and N.~L. Zhang.
\newblock Hidden-mode markov decision processes.
\newblock In {\em Proceedings of the 16th international joint conference on
  artificial intelligence (IJCAI-99), workshop on neural symbolic, and
  reinforcement methods for sequence learning, Stockholm, Sweden}, 1999.

\bibitem{dragan2013policy}
A.~D. Dragan and S.~S. Srinivasa.
\newblock A policy-blending formalism for shared control.
\newblock {\em The International Journal of Robotics Research}, 32(7):790--805,
  2013.

\bibitem{feldman2010model}
A.~Feldman, G.~Provan, and A.~Van~Gemund.
\newblock A model-based active testing approach to sequential diagnosis.
\newblock {\em Journal of Artificial Intelligence Research}, 39:301--334, 2010.

\bibitem{flad2017cooperative}
M.~Flad, L.~Fr{\"o}hlich, and S.~Hohmann.
\newblock Cooperative shared control driver assistance systems based on motion
  primitives and differential games.
\newblock {\em IEEE Transactions on Human-Machine Systems}, 47(5):711--722,
  2017.

\bibitem{iosti2020synthesizing}
S.~Iosti, D.~Peled, K.~Aharon, S.~Bensalem, and Y.~Goldberg.
\newblock Synthesizing control for a system with black box environment, based
  on deep learning.
\newblock In {\em International Symposium on Leveraging Applications of Formal
  Methods}, pages 457--472. Springer, 2020.

\bibitem{jain2019probabilistic}
S.~Jain and B.~Argall.
\newblock Probabilistic human intent recognition for shared autonomy in
  assistive robotics.
\newblock {\em ACM Transactions on Human-Robot Interaction (THRI)}, 9(1):1--23,
  2019.

\bibitem{javdani2018shared}
S.~Javdani, H.~Admoni, S.~Pellegrinelli, S.~S. Srinivasa, and J.~A. Bagnell.
\newblock Shared autonomy via hindsight optimization for teleoperation and
  teaming.
\newblock {\em The International Journal of Robotics Research}, 37(7):717--742,
  2018.

\bibitem{jiang2021goal}
Y.-S. Jiang, G.~Warnell, and P.~Stone.
\newblock Goal blending for responsive shared autonomy in a navigating vehicle.
\newblock In {\em The AAAI Conference on Artificial Intelligence}, volume~35,
  pages 5939--5947, 2021.

\bibitem{losey2018review}
D.~P. Losey, C.~G. McDonald, E.~Battaglia, and M.~K. O'Malley.
\newblock A review of intent detection, arbitration, and communication aspects
  of shared control for physical human--robot interaction.
\newblock {\em Applied Mechanics Reviews}, 70(1):010804, 2018.

\bibitem{marcano2020review}
M.~Marcano, S.~D{\'\i}az, J.~P{\'e}rez, and E.~Irigoyen.
\newblock A review of shared control for automated vehicles: Theory and
  applications.
\newblock {\em IEEE Transactions on Human-Machine Systems}, 50(6):475--491,
  2020.

\bibitem{mirsky2016sequential}
R.~Mirsky, R.~Stern, M.~Kalech, et~al.
\newblock Sequential plan recognition.
\newblock In {\em IJCAI International Joint Conference on Artificial
  Intelligence}, volume 2016, pages 401--407, 2016.

\bibitem{olsson2009literature}
F.~Olsson.
\newblock A literature survey of active machine learning in the context of
  natural language processing.
\newblock 2009.

\bibitem{peled2019control}
D.~Peled, S.~Iosti, and S.~Bensalem.
\newblock Control synthesis through deep learning.
\newblock In {\em From Reactive Systems to Cyber-Physical Systems}, pages
  242--255. Springer, 2019.

\bibitem{peternel2018robot}
L.~Peternel, N.~Tsagarakis, D.~Caldwell, and A.~Ajoudani.
\newblock Robot adaptation to human physical fatigue in human--robot
  co-manipulation.
\newblock {\em Autonomous Robots}, 42:1011--1021, 2018.

\bibitem{philips2007adaptive}
J.~Philips, J.~d.~R. Mill{\'a}n, G.~Vanacker, E.~Lew, F.~Gal{\'a}n, P.~W.
  Ferrez, H.~Van~Brussel, and M.~Nuttin.
\newblock Adaptive shared control of a brain-actuated simulated wheelchair.
\newblock In {\em 2007 IEEE 10th International Conference on Rehabilitation
  Robotics}, pages 408--414. IEEE, 2007.

\bibitem{selvaggio2019haptic}
M.~Selvaggio, R.~Moccia, F.~Ficuciello, B.~Siciliano, et~al.
\newblock Haptic-guided shared control for needle grasping optimization in
  minimally invasive robotic surgery.
\newblock In {\em 2019 IEEE/RSJ International Conference on Intelligent Robots
  and Systems (IROS)}, pages 3617--3623. IEEE, 2019.

\bibitem{settles2012active}
B.~Settles.
\newblock Active learning.
\newblock {\em Synthesis lectures on artificial intelligence and machine
  learning}, 6(1):1--114, 2012.

\bibitem{shannon1948mathematical}
C.~E. Shannon.
\newblock A mathematical theory of communication.
\newblock {\em The Bell system technical journal}, 27(3):379--423, 1948.

\bibitem{shvo2020active}
M.~Shvo and S.~A. McIlraith.
\newblock Active goal recognition.
\newblock In {\em Proceedings of the AAAI Conference on Artificial
  Intelligence}, volume~34, pages 9957--9966, 2020.

\bibitem{siddiqi2011sequential}
S.~A. Siddiqi and J.~Huang.
\newblock Sequential diagnosis by abstraction.
\newblock {\em Journal of Artificial Intelligence Research}, 41:329--365, 2011.

\bibitem{sutton2018reinforcement}
R.~S. Sutton and A.~G. Barto.
\newblock {\em Reinforcement learning: An introduction}.
\newblock MIT press, 2018.

\bibitem{wu2015improving}
Y.~Wu, W.~L. Chan, Y.~Li, K.~P. Tee, R.~Yan, and D.~K. Limbu.
\newblock Improving human-robot interactivity for tele-operated industrial and
  service robot applications.
\newblock In {\em 2015 IEEE 7th International Conference on Cybernetics and
  Intelligent Systems and IEEE Conference on Robotics, Automation and
  Mechatronics}, pages 153--158. IEEE, 2015.

\bibitem{zeng2020view}
R.~Zeng, Y.~Wen, W.~Zhao, and Y.-J. Liu.
\newblock View planning in robot active vision: A survey of systems,
  algorithms, and applications.
\newblock {\em Computational Visual Media}, 6(3):225--245, 2020.

\end{thebibliography}
